**Deep neural networks with controlled variable selection for the identification of putative causal genetic variants**

Peyman H. Kassani[1], Fred Lu[2], Yann Le Guen[1], Zihuai He[1,3#]

[1]Department of Neurology and Neurological Sciences, Stanford University, Stanford, CA, USA

[2]Department of Statistics, Stanford University, Stanford, CA, USA

[3]Quantitative Sciences Unit, Department of Medicine (Biomedical Informatics Research), Stanford University, Stanford, CA, USA

# Correspondence to: Zihuai He (zihuai@stanford.edu)

**Abstract**

Deep neural networks (DNN) have been used successfully in many scientific problems for their high prediction accuracy, but their application to genetic studies remains challenging due to their poor interpretability. In this paper, we consider the problem of scalable, robust variable selection in DNN for the identification of putative causal genetic variants in genome sequencing studies. We identified a pronounced randomness in feature selection in DNN due to its stochastic nature, which may hinder interpretability and give rise to misleading results. We propose an interpretable neural network model, stabilized using ensembling, with controlled variable selection for genetic studies. The merit of the proposed method includes: (1) flexible modelling of the non-linear effect of genetic variants to improve statistical power; (2) multiple knockoffs in the input layer to rigorously control false discovery rate; (3) hierarchical layers to substantially reduce the number of weight parameters and activations to improve computational efficiency; (4) de-randomized feature selection to stabilize identified signals. We evaluated the proposed method in extensive simulation studies and applied it to the analysis of Alzheimer's disease genetics. We showed that the proposed method, when compared to conventional linear and nonlinear methods, can lead to substantially more discoveries.

**Introduction**

Recent advances in whole genome sequencing (WGS) technology have led the way to explore the contribution of common and rare variants in both coding and non-coding regions towards risk for complex traits. Large-scale genome sequencing studies, such as the Trans-Omics for Precision Medicine (TOPMed) study and the Alzheimer's Disease Sequencing Project (ADSP), have collected thousands of samples with directly sequenced whole genomes. One main theme in WGS studies is to understand the genetic architecture of disease phenotypes and, critically, provide a credible set of well-defined, novel targets for the development of genomic-driven medicine[1]. However, the identification of causal variants in these datasets remains challenging due to the sheer number of genetic variants, low signal-to-noise ratio, and strong correlations among genetic variants. Most of the results published so far are derived from marginal association models that regress an outcome of interest on the linear effect of a single genetic variant or multiple genetic variants in a gene[2,3]. The marginal association is quantified by a p-value. Genetic variants or genes below a p-value threshold are deemed as associated variants. The marginal association tests are well-known for their simplicity and effectiveness, but they often identify proxy variants that are only correlated with the true causal variants, and the statistical power can be suboptimal.

Deep neural networks (DNNs) can efficiently learn the nonlinear effects of data on an outcome of interest using hidden layers in its framework[4] without any explicit assumptions. DNNs have gained popularity for their superior performance in many scientific problems, showing exceptional prediction accuracy in many domains, including image studies such as object detection, recognition, and segmentation[5–7]. Although there are now large-scale genetic data available to potentially embark on deep learning approaches to genetic data analysis, the applications of DNN methods to WGS studies have been limited. One obstacle for the widespread application of DNN to genetic data is their interpretability. Unlike linear regression or logistic regression, it is generally difficult to identify how changes of the genetic variants influence the disease outcome in a DNN because of its multilayer nonlinear structure. While several methods have been developed to improve the interpretability of neural networks and quantify the relative importance of input features, most methods lack a rigorous control on the false discovery rate (FDR) of selected features[8]. Moreover, existing methods are susceptible to noise and a lack of robustness. Ghorbani et al[9]. have shown that small perturbations can change the feature importance and can lead to dramatically different interpretations given the same dataset. In this paper, we consider the scalable, robust variable selection problem in DNN for the identification of putative causal genetic variants in WGS studies.

Knockoff framework[10] is a recent breakthrough in statistics that is designed for feature selection with rigorous FDR control. The idea of knockoff-based inference is to generate synthetic, noisy copies (knockoffs) of the original features where each sample resembles the original data in terms of the feature correlation structure but is conditionally independent of the trait of interest, given the original features. In a learning model with both original features and their knockoff counterparts, the knockoffs serve as negative controls for feature selection. The Model-X knockoffs[11] does not require any relationship linking genotypes to phenotypes and imposes no restriction on the number of features relevant to the sample size. Therefore, it can naturally bring interpretability to any learning method, including but not limited to marginal association tests, joint linear models such as Lasso[12], and nonlinear DNN. Notably, Sesia et al.[13] proposed *KnockoffZoom* for genetic studies based on a linear Lasso regression[12]. For feature selection in nonlinear DNN, Lu et al[14]. proposed *DeepPINK* based on knockoffs and Song and Li[8] proposed *SurvNet* using conceptually similar surrogate variables, but they are not scalable to current genetic studies; moreover, the knockoff copies / surrogate variables are randomly generated, adding extra randomness to the interpretation of DNN that is already fragile. He et al.[15] proposed *KnockoffScreen* to utilize multiple knockoffs to improve the stability of feature selection, but it is built upon conventional marginal association tests in genetic studies which do not account for nonlinear and interactive effects of genetic variants.

In this paper, we couple hierarchical DNN with multiple knockoffs to develop an interpretable DNN for identification of putative causal variants in genome sequencing studies with rigorous FDR control. Besides the modelling of the non-linear effects for enhanced power and the use of multiple knockoffs for improved stability, there are three additional contributions of this paper: first, we propose a hierarchical DNN architecture that is suitable for the analysis of common and rare variants in sequencing studies, which also allows adjustment of potential confounders. The new architecture requires orders of magnitude fewer parameters and activations compared to a fully connected neural network model. Second, we identified a vanishing gradient problem[16] due to the presence of low-frequency and rare variants in sequencing studies, and propose a practical solution using the exponential rectified linear unit (ELU) activation function[17]. This modification leads to substantial gain in power compared to the popular rectified linear unit (ReLU) activation function[18]. Third, we identified a pronounced randomness in feature selection in DNN due to its stochastic nature, which may hinder the interpretability and give rise to misleading results. We observed that different runs of a DNN with identical hyper-parameters produce inconsistent feature importance scores, although the difference in prediction accuracy is negligible. We propose a de-randomized feature selection that enables robust interpretation of DNN. We applied the proposed method to the analysis of Alzheimer's disease genetics using data from 10,797 clinically diagnosed AD cases and 10,308 healthy controls.

**Results**

**Overview of the proposed hierarchical deep learning with multiple knockoffs (HiDe-MK)**

The workflow summary of the proposed Hide-MK is presented in **Figure 1**. The aim is to develop a deep learning-based variable selection method with FDR control guarantees through the knockoff framework. For each genetic variant (feature), we first construct five knockoff variables that are simultaneously exchangeable to the original feature but conditionally independent of the disease outcome (**Figure 1a**). The generation of multiple knockoffs for genetic data is based on the sequential conditional independent tuples (SCIT) algorithm proposed by He et al.[15]. Both original cohort and synthetic cohorts are fed into the neural network as inputs. Knockoffs are served as control features during training and thereafter help tease apart the true signals which are explanatory for the response variable $y \in \mathbb{R}^n$, where $n$ is the sample size.

The proposed neural network includes two hierarchical layers (**Figure 1b**), which are locally connected dense layers. The first layer of hierarchy concatenates each original feature and its multiple knockoffs as the input for each neuron. The second layer of hierarchy concatenates adjacent genetic variants in a nearby region. This was inspired by recent advances for the gene/window-based analysis of WGS data, where multiple common and rare variants are grouped for improved power[19,20]. The output of the second hierarchy includes multiple channels (filters) which help maximally learn the information of a local region and exploit the local correlation in each group. These two layers of hierarchy substantially reduce the size of the parameter space compared to using standard dense layers. Then, the resulting neurons are fed to dense layers and then linked to the output layer together with additional covariates such as age, gender and principal components that are used as controls for population stratifications (**Figure 1b**, neurons shown by green color).

The next step is to obtain feature importance scores (FIs), i.e., the importance of each genetic variant. The influence of feature $\mathbf{x}_i$, $i = 1, \ldots, p$ to the response $y \in \mathbb{R}^n$ is measured[21] via the gradient for both true and knockoff features throughout forward and backward passes (**Figure 1c**). The gradient information is then summarized as FIs. Details on FIs' calculation are discussed in Methods section. Hyper-parameters of HiDe-MK were tuned based on a five-fold cross-validation. With optimal hyper-parameters and epoch number, we re-fitted the model to the whole data and calculated FIs. Due to the random nature of fitting neural network models and the identifiability issue due to the large number of weight parameters, we repeatedly fit the model ten times, given the same data and with identical hyper-parameters, and then took the median of FIs over ten runs. The median naturally brings robustness to outlier FIs. We also observed that a DNN at different epochs can produce inconsistent feature importance scores, although the difference in prediction accuracy is negligible. Thus we slightly modified the criteria to choose optimal epoch number to stabilize the feature selection. We demonstrated in both simulation studies and real data analysis that this de-randomized approach, referred to as de-randomized HiDe-MK, substantially reduced variability and improved the stability of FIs.

Once FIs for original and knockoff features were obtained, a knockoff filter was applied to select causal features with controlled errors at different target FDR threshold values (e.g., 0.10, 0.20) (**Figure 1d**). We used the knockoff filter for multiple knockoffs proposed in He et al.[15], which leverages multiple knockoffs for improved power, stability and reproducibility. In the Methods section, we describe in detail the knockoff generation, network specifications (activation functions, hyper-parameters, regularizations, etc.), feature importance calculation, de-randomization and feature selection.

**Figure 1: Overview of the workflow.** The four panels show **a**) the knockoff feature generation. In this study, we generated 5 sets of knockoffs using SCIT; **b**) The proposed hierarchical deep learner. We utilized two layers of hierarchy to substantially decrease the size of the network. We also used ELU activation function for better network performance; **c**) Gradients were used to measure feature importance scores. Each instance of data flows throughout the network and its gradient is monitored. The size of FIs is same as input data; **d**) obtained FIs were used to select causal putative genetic variants via knockoff filter.

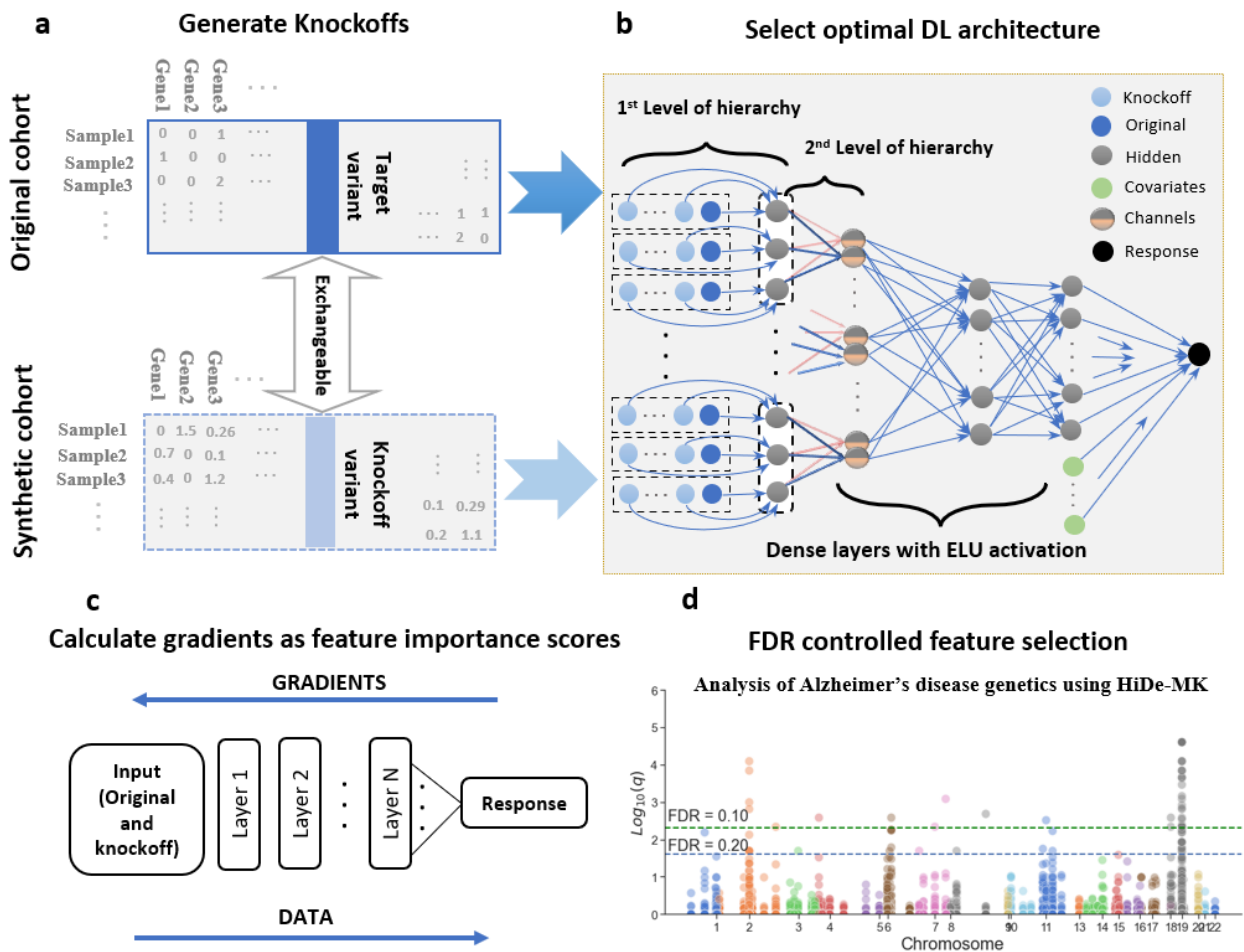

**De-randomized HiDe-MK improved power with FDR control guarantees in simulation studies.**

We carried out simulation studies for both quantitative and dichotomous outcomes (regression and classification tasks). The aim is to evaluate the FDR and power of the proposed *de-randomized HiDe-MK* compared to the proposed *HiDe-MK* and other conventional methods including support vector machines for classification (SVM) and regression (SVR), multilayer perceptron (MLP), least absolute shrinkage and selection operator (Lasso), ridge regression, DeepPINK, our modified version (Modif-DeepPINK), in which an 'ELU' activation function is used instead of ReLU, and a marginal association test quantified by p-values (Marginal-test) for both quantitative and dichotomous traits. We used the R package GLMNET[22] to implement Lasso and Ridge regressions and R package LibLineaR[23] to implement SVR and SVM. The marginal test was based on a Wald's test in generalized linear models. For a fair comparison, all methods are based on knockoff inference that controls FDR. Different methods represent different calculations of feature importance score. The proposed HiDe-MK is equipped with five knockoffs and is compared with a deep learner with multiple knockoffs (De-MK) which has one layer of hierarchy less than HiDe-MK. Besides *de-randomized HiDe-MK, HiDe-MK* and *De-MK*, other methods are equipped with a single knockoff similar to DeepPINK. Knockoffs are generated by the SCIT method proposed in KnockoffScreen[15]. For simulating the sequence data, each replicate consists of 10,000 individuals with genetic data on 2,000 genetic variants from a 200 kb region, simulated using the haplotype dataset in the SKAT package. The SKAT haplotype dataset was generated using a coalescent model (COSI), mimicking the linkage disequilibrium structure of European ancestry samples. We restrict the simulation studies to genetic variants with minor allele count (MAC) greater than 10, resulting in 400-500 genetic variants as input features. The quantitative and dichotomous outcomes are simulated as a non-linear function of the genetic variants. Simulation details are described in the Methods section.

For each replicate, the empirical power is defined as the proportion of detected true signals among all causal signals and the empirical FDR is defined as the proportion of false signals among all detected signals. Based on 200 replicates, we report the average empirical power and observed FDR at different levels of target FDR from 0.01 to 0.20 with step size 0.01. Results are depicted in **Figure 2**. The proposed method exhibits substantially higher power at low target FDR (e.g. target FDR=0.05; for both panels of **Figure 2**) compared to a single knockoff-based method which is limited by the detection threshold as discussed in Gimenez and Zou (2018)[24]. In addition, we observed a higher power than other linear alternatives at high target FDR (e.g., 0.2) because a DNN is able to account for nonlinear effects over genetic variants. Referring to **Figure 2**, a modified DeepPINK gains better power than its conventional one in both scenarios. We found that DeepPINK with ReLU activation function suffers from the vanishing gradients associated with the "dying ReLU neurons" problem[18]. De-MK shows similar power as HiDe-MK; however De-MK is much denser and uses more weight parameters than a HiDe-MK (refer to **Figure 5** for an illustration). Among linear models, Lasso and Ridge were the best models for dichotomous and quantitative traits. Marginal test was also very competitive and was the second-best model among linear models, for quantitative trait, but it showed poor power for dichotomous trait. It is worth noting that the prediction accuracy of HiDe-MK is comparable to Lasso (e.g., for dichotomous trait, we observed 0.01 difference in terms of validation AUC in favor of HiDe-MK), although the power difference is substantial. This demonstrates that the usual criteria for prediction cannot be directly applied to feature selection.

**Figure 2: Power and false discovery rate (FDR) comparison.** Two panels show observed power and FDR for dichotomous and quantitative traits with varying target FDR from 0.01 to 0.20. Different colors indicate different learning method. De-randomized HiDe-MK: derandomized version of hierarchical deep neuro-network with multiple knockoffs; HiDe-MK: hierarchical deep neural network with multiple knockoffs; De-MK: deep neural network with multiple knockoffs, Modif-DeepPINK: a modified variant of DeepPINK, DeepPINK: deep feature selection using paired-input nonlinear knockoffs), MLP: multilayer perceptron, SVM: support vector machines for classification, SVR: support vector machines for regression, Lasso: least absolute shrinkage and selection operator, Ridge: ridge regression, Marginal test: marginal association test quantified by p-values, and equipped with a single set of knockoff feature.

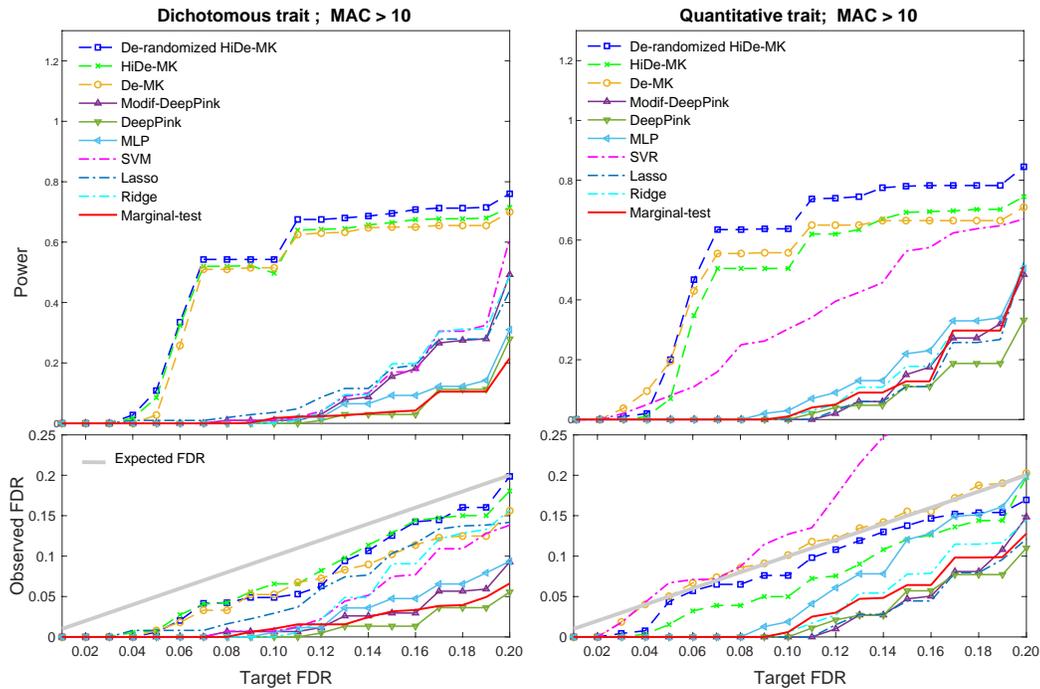

## Application of de-randomized HiDe-MK to GWAS

The main goal of WGS studies is to identify genetic variants associated with certain disease phenotypes, referred to as genome-wide association studies (GWAS). The task is commonly carried out in multiple stages, including an exploratory stage to discover candidate regions/loci, and a confirmatory stage to pin down the final discoveries[25]. As a proof of concept, we applied the linear method *KnockoffScreen* to the analysis of Alzheimer's disease (AD) in 388,051 WG-imputed samples from the UK Biobank to discover candidate regions/loci, and then applied de-randomized HiDe-MK to the confirmatory stage using an independent cohort of 10,797 clinically diagnosed AD cases and 10,308 healthy controls. The candidate regions in the confirmatory stage includes 472 variants associated with Alzheimer's disease (394 from UKB analysis; 78 from previous GWAS) and with a 5kb window centered on each locus. The final dataset for the confirmatory stage includes 21,105 samples with 11,662 genetic variants. Five knockoff variants were generated for each variant. The total number of input features for de-randomized HiDe-MK is $11{,}662 \times 6 = 69{,}972$. We additionally adjusted the learner for sex and 10 leading principal components as covariates. Details for dataset preparation is explained in Methods section.

We report results based on target FDRs 0.10 and 0.20. Each identified genetic variant is then assigned to either a gene or an intergenic region. If the variant is within a gene, we report the gene's name and if it is in an intergenic region, we report the upstream and downstream genes. For comparison, we considered feature selection in Lasso with a single knockoff, similar to *KnockoffZoom*[13]. We present the results in **Table 1** and **Figure 3**. We observed that de-randomized HiDe-MK identified 72 AD-associated genetic variants that pass the target FDR 0.20, corresponding to 20 proximal genes (**Table 1**). By contrast, Lasso is able to identify only seven AD-associated variants corresponding to six AD-associated genes at target FDR 0.20. In addition, Lasso cannot detect any genes at FDR threshold 0.10 whereas de-randomized HiDe-MK identified 44 variants corresponding to 14 genes that pass FDR threshold 0.10 (**Figure 3**). More details for the position of each identified SNP, rsIDs, and proximal genes are listed in **Table S1**.

**Table 1. List of identified genes for both Lasso with a single knockoff and de-randomized HiDe-MK**.

|  | **Identified Genes** |
|---|---|
| **Lasso** | *AC011481.3, APOE, CLASRP, TREM2, ABI3, MARK4* |
| **De-randomized HiDe-MK** | *BIN1, SLMAP, CD2AP, GPC2, GAL3ST4, EPHA1, EPHA1-AS1, SHARPIN, PICALM, SPI1, SLC24A4, ABCA7, BCL3, NECTIN2, AC011481.2, AC011481.3, TOMM40, APOE, APOC1, CLPTM1* |

**Figure 3: Manhattan plots for a de-randomized HiDe-MK (top) and a Lasso (bottom)**. Each dot point represents a genetic variant. The dashed horizontal lines indicate target FDR 0.10 and 0.20.

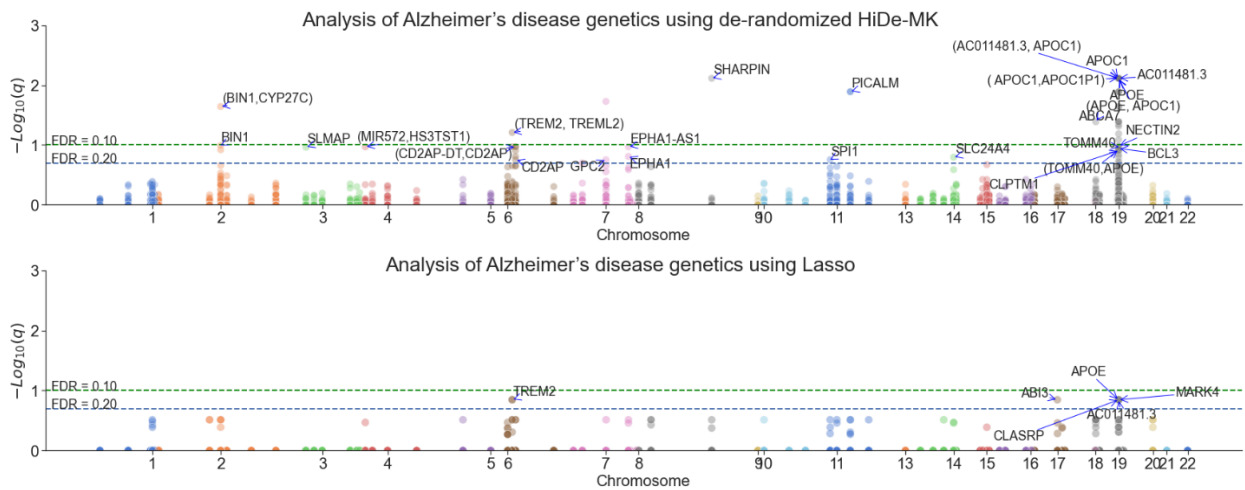

**Feature importance stability analysis: de-randomization helps stabilize HiDe-MK**

In the application of HiDe-MK to the analysis of Alzheimer's disease genetics, we observed that different runs of HiDe-MK with same dataset, knockoff features, hyper-parameters, epoch number and validation loss can lead to drastically different feature importance score and therefore produce different set of selected features. This is due to the non-convexity of deep learning methods which rely on random parameter initialization and stochastic gradient descent to reach a local optimum. The resulting gradient-based FIs thus tend to be stochastic. We present the randomness of feature importance values (FIs) in

**Figure 4** (left panel), in terms of the correlation between FIs across ten runs of HiDe-MK. Each time HiDe-MK was applied to the aforementioned AD genetics data, with identical knockoff features, hyper-parameters and epoch number. Although the difference in validation loss is negligible, we observed a poor correlation (<0.60) between different runs. The result shows that direct application of conventional DNN can give rise to misleading results. It also implies that the usual criteria for prediction cannot be directly applied to feature selection.

To have a DNN method reliably express the relationship between genotype and phenotype, the neural network and its feature importance values should be de-randomized, and the criteria to choose the optimal epoch number should be modified. First, we propose using an ensemble of 10 runs of HiDe-MK with identical and optimal set of hyper-parameters. We then take the median of FIs over the 10 runs for each feature. Second, we modified the rule to choose the optimal epoch number as detailed in the method section. We refer to this ensemble method as de-randomized HiDe-MK. We present the empirical results in **Figure 4** (middle panel). We observed a high correlation (>0.85) between different runs of de-randomized HiDe-MK, where each run ensembles 10 sets of FIs, demonstrating that the ensemble method helps de-randomize FIs. This step of de-randomization was crucial in our modelling since our goal was to report a credible set of AD-associated genes. We also present the results for the same approach taking the mean instead of median in **Figure S4**. In addition, we evaluated the stability of FIs across epoch numbers. For every epoch, we monitored the knockoff feature statistics W and calculated the correlation for every two consecutive epochs. We evaluate both HiDe-MK and its derandomized version (**Figure 4,** right panel). We observed a high correlation in FIs between two consecutive epochs as epoch number increases. We also observed that the de-randomized HiDe-MK is more stable than HiDe-MK.

Beside the randomness due to the stochastic nature of deep learning methods, many interpretable DNNs rely on a set of randomly generated "control" features, such as the surrogate variables in *SurvNet* and the knockoff variables in the proposed methods. We also note that randomness brought by surrogate/knockoff variables may also hinder the interpretability. We propose using multiple knockoffs and a corresponding knockoff filter to stabilize the feature selection. Detailed comparison with single knockoff methods was discussed by He et al. (2021)[15].

**Figure 4: De-randomized HiDe-MK demonstrates low randomness compared to a single run of HiDe-MK**. The correlation matrix of ten different runs of HiDe-MK with identical learning hyper-parameters reveals a drastic randomness in its every run (Left). Taking median of feature importance values (FIs) for every ten runs substantially de-randomized FIs (Middle). We also measured the stability of FIs across epoch numbers in terms of knockoff feature statistics W for both HiDe-MK and de-randomized HiDe-MK (Right). Its de-randomized variant is obviously more stable.

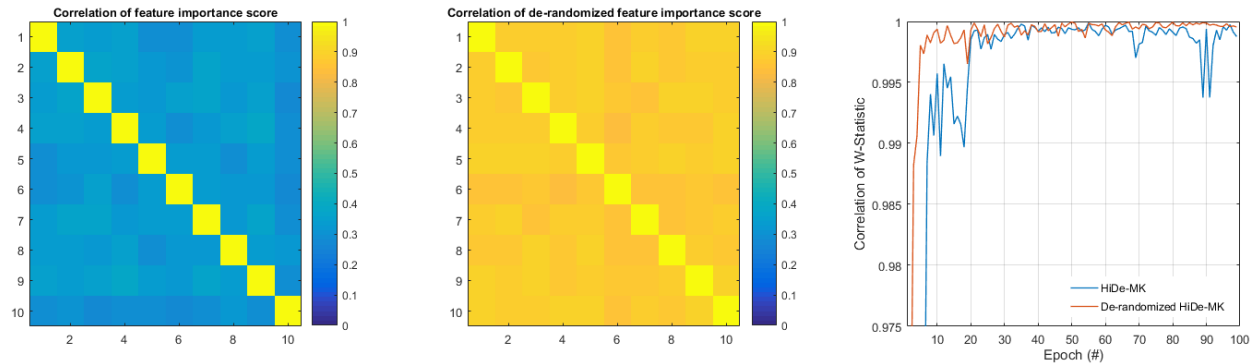

**The hierarchical architecture improves computational efficiency**

The computational cost plays a key role in the application of deep learning methods to genetics data in the presence of knockoffs data which multiplies the number of input features. Deep learning methods which use several hidden layers in their structure can be computationally intensive. As the number of features increases, the network size in terms of the number of weight parameters and number of activations gets larger and consequently the computational burden increases. To illustrate the importance of hierarchical structure in DNN, we compared three different learning methods: fully connected neural network (FCNN), neural network with one level of hierarchy (1-level of hierarchy) and neural network with two levels of hierarchy (2-level of hierarchy). FCNN is a conventional DNN with three dense hidden layers.. The 1-level of hierarchy model uses an initial locally connected layer to join each original feature with its five knockoffs from input layer to the next layer, and this reduced set of neurons is connected to the remaining layers with the same architecture as FCNN. The 2-level of hierarchy model uses one more level of hierarchy compared to a 1-level hierarchy, to group adjacent genetic variants, replacing the corresponding dense layer. Batch size, and the number of epochs are set to 1024, and 50 respectively.

We applied these methods to the final genetic data consisting of 21,105 individuals and 11,662 genetic variants. On our computing system, we noticed that experiments with FCNN causes out-of-memory error due to the huge number of its weight parameters. Therefore, we limited our experiments to the random selection of only 1000 genetic variants and their five set of knockoffs as a proof of concept. We ran these models for 50 times and reported the average number of weights, computational time and the number of activations. Results are demonstrated in **Figure 5**. 2-level of hierarchy has two orders of magnitude fewer weight parameters in its architecture than a 1-level hierarchy and four orders of magnitude fewer than a FCNN which does not use any hierarchical layer. The middle panel displays the averaged time per epoch for three counterparts. 2-level hierarchy is 2 times faster than 1-level hierarchy and 40 times faster than a FCNN. We also quantified the number of activations as it is also an important factor in measuring the model's efficiency[26]. Again, 2-level hierarchy uses about 2 and 12 times fewer activation functions than 1-level hierarchy and FCNN respectively. DNNs with hierarchical layers are significantly more efficient than DNNs with dense layers.

**Figure 5: Experiments with DNN variants equipped with five set of knockoffs**. The comparison was measured on data with 21,105 individuals, 1000 randomly selected genetic variants, and five knockoffs per variant. The left panel presents the number of weights of the three models. The middle panel presents the averaged time per epoch. The right panel presents the number of activation functions.

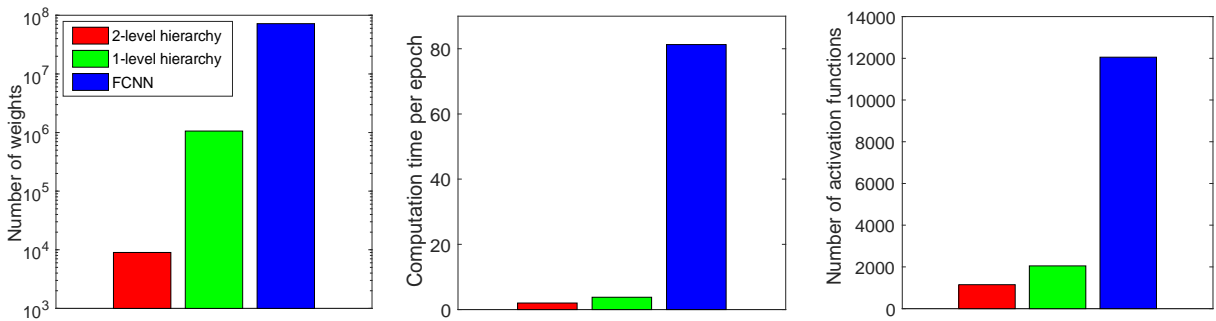

## Conclusion

In this study, we proposed an interpretable DNN denoted as de-randomized HiDe-MK for the identification of putative causal genetic variants in WGS studies. We took advantage of the localizable structure of genetic variants through hierarchical layers in the architecture of DNN in order to seamlessly reduce the size of the DNN. We further employed knockoff framework with multiple set of knockoffs to rigorously control FDR during feature selection. While the underlying goal is to identify putative causal variants through an exploratory stage, we observed a nontrivial randomness in the selected genetic variants. Two different runs of any DNNs including HiDe-MK from the same hyper-parameters led to different candidate variants which we found concerning. To stabilize identified signals, we proposed an ensemble method aggregating the feature importance score from ten "identical" runs, which allows us to confidently determine the final selected features with less variance. With a thorough experiment conducted on two simulation datasets, validated for both regression and classification, on both common and rare variants, we empirically showed this de-randomization stage improves empirical power with a controlled FDR and substantially increases the feature importance stability. For real data analysis, we applied de-randomized HiDe-MK to the confirmatory stage of a GWAS. De-randomized HiDe-MK confirmed several genetic variants previously reported that were missed by a linear Lasso. This may shed light on the discovery of additional risk variants using sophisticated DNNs in future genetic studies.

We believe this study can be seen as a foundation, while emphasizing potential pitfalls, for other interpretable DNN methods. We observed the prediction accuracy by itself is inadequate as a single criterion for model training if the goal is feature selection. Optimal hyper-parameters and causal variants should be regarded and selected cautiously. Small changes in hyper-parameters can lead to different sets of identified variants while changes in prediction accuracy is negligible. A de-randomization step is necessary in designing interpretable DNNs. Future research should aim for further improving the DNN robustness to its stochastic nature, small perturbations in the input features, and requirement of hyperparameter tuning.

## Methods

### Simulated data to evaluate empirical FDR and power

An extensive experiment is performed to evaluate the empirical FDR and power. The initial size of genetics data for doing simulation consists of 10,000 individuals with 2,000 genetic variants drawn from a 200kb region, based on a coalescent model (COSI) mimicking LD structure of European ancestry[27]. Simulations were devised for both rare and common variants with minor allele content (MAC) >10. We followed the settings in KnockoffScreen[15] with slight modification. Strong correlation among variants (known as tightly linked variants) may make it difficult for learning methods to distinguish a causal genetic variant from its highly correlated counterpart (see Sesia et al[28]). Therefore, we only pick variants from each tightly linked cluster in the presence of strong correlations. Specifically, a hierarchical clustering is first applied to variants to not allow two clusters to share cross-correlations greater than 0.75 and then variants from each cluster are randomly chosen as candidates and are included in our simulation studies[15]. We set four variants in a 200kb region as causal variants. We evaluated quantitative and dichotomous traits generated by:

$$\text{Quantitative trait: } Y_i = X_{i1} + Cf(\beta_1 g_1 + \cdots + \beta_s g_s) + \varepsilon_i,$$

$$\text{Dichotomous trait: } g(\mu_i) = \beta_0 + X_{i1} + Cf(\beta_1 g_1 + \cdots + \beta_s g_s)$$

where $X_{i1} \sim N(0,1)$, $\varepsilon_i \sim N(0,2)$ and they are all independent; $(g_1, \ldots, g_s)$ are selected risk variants; $g(x) = \log x / \log(1 - x)$ and $\mu_i$ is the conditional mean of the $i$-th target. For the dichotomous trait, $\beta_0$ is chosen by the prevalence 0.10. The effect $\beta_j = \frac{a}{\sqrt{2m_j(1-m_j)}}$, where $m_j$ is the minor allele frequency (MAF) for the $j$-th variant. Parameter $a$ is defined such that the $\beta_1^2 var(g_1) + \cdots + \beta_s^2 var(g_s) = 0.2$ for dichotomous trait and 0.06 for the quantitative trait. To mimic the real data scenario with both risk variants and protective variants, we set $\beta_1$ to be negative and the others to be positive. Additionally, we considered non-linear effect of the causal variants. We define $f(x) = x^2$ and $C = 2$ for both traits. With this setup, the number of genetic variants which includes both common and rare variants is in a range between 400 to 500. We generate 200 replicates for each trait, and we report the average FDR and average power at different target FDR threshold values. We applied several machine learning methods and illustrated results in two panels in **Figure 2**.

**Confirmatory stage genetic data**

We queried 45,212 individual genotypes from 28 cohorts genotyped on genome-wide microarrays and imputed at high resolution based on the reference panels from TOPMed using the Michigan Imputation Server[29]. We restricted our analysis to European ancestry individuals. After quality control, restricting to case/control status, pruning for duplicates, and the 3rd degree relatedness, 21,105 unique individuals remained for the analysis. We considered 78 candidate variants from previous GWAS studies listed in Andrews et. al.[30] and 394 candidate regions identified by a UK Biobank analysis using KnockoffScreen. We additionally included genetic variants in the 5kb of the neighborhood of each candidate variant/region. The final dataset consists of 21,105 subjects and 11,662 variants. Samples used in this manuscript are derived from the replication set imputed on the TOPMed reference panel and described in Le Guen et al.[31]

**Overview of the knockoff framework**

Controlling the FDR when performing variable selection can be accomplished by the knockoff framework. With this purpose, a set of variants, so called "knockoffs" denoted by $\widetilde{\mathbf{X}} \in \mathbb{R}^{n \times p}$ with the same size of the original input $\mathbf{X} \in \mathbb{R}^{n \times p}$ should be created where $p$ accounts for genetic variants and $n$ for the total number of individuals. Since knockoffs are conditionally independent of response vector $\mathbf{y} \in \mathbb{R}^n$, we expect those true variables to exhibit higher association with $\mathbf{y} \in \mathbb{R}^n$ than their knockoff counterparts. The knockoffs framework can be summarized in four steps:

(1) Generate multiple knockoffs for each true variant.

(2) Calculate the feature importance scores (FIs) for the original variants and the knockoff variants; FIs are assigned by a data-driven learning model.

(3) Calculate the feature statistic by contrasting FIs between the original and their knockoff counterparts.

(4) Apply knockoff filter to select significant variants with FDR control.

In the following we explain these steps.

Generate multiple knockoff variables

To generate knockoff variants $\widetilde{\mathbf{X}} \in \mathbb{R}^{n \times p}$ two properties should be deemed: 1) $\widetilde{\mathbf{X}} \in \mathbb{R}^{n \times p}$ is independent of $\mathbf{y} \in \mathbb{R}^n$ conditional on $\mathbf{X} \in \mathbb{R}^{n \times p}$; 2) $\mathbf{X}$ and $\widetilde{\mathbf{X}}$ are exchangeable[11]. With this setup, knockoff variants can serve as control variables for feature selection. There are two limitations for a single knockoff procedure; 1) a single knockoff is limited by the detection threshold $[\frac{1}{\alpha}]$, which is the minimum number of independent rejections that are required in order to detect any association[24]; 2) a single knockoff is not stable in terms of the selected sets of features; i.e., two different runs of a single knockoff may generate different sets of features and lead to different selected features. To reduce the randomness issue and improve power, we used the efficient sequential conditional independent tuples (SCIT) algorithm proposed by He et. al.[15] to generate multiple knockoffs that are simultaneously exchangeable. Algorithm 1 shows the main steps of SCIT which yield a sequence of random variables obeying the exchangeability property.

**Algorithm 1** Sequential Conditional Independent Tuples (Multiple Knockoffs)

$j = 1$
**while** $j \leq p$ **do**
  Sample $\tilde{\mathbf{x}}_j^1, \cdots, \tilde{\mathbf{x}}_j^M$ independently from $\mathcal{L}(\mathbf{x}_j | \mathbf{x}_{-j}, \tilde{\mathbf{x}}_{1:j-1}^1, \cdots, \tilde{\mathbf{x}}_{1:j-1}^M)$
  $j = j + 1$
**End**

where $\mathcal{L}(\mathbf{x}_j | \mathbf{x}_{-j}, \tilde{\mathbf{x}}_{1:j-1}^1, \cdots, \tilde{\mathbf{x}}_{1:j-1}^M)$ is the conditional distribution of $\mathbf{x}_j$ given $\mathbf{x}_{-j}, \tilde{\mathbf{x}}_{1:j-1}^1, \cdots, \tilde{\mathbf{x}}_{1:j-1}^M$ in which '-j' means the variable $\mathbf{x}_j$ is excluded and $M$ is the total number of knockoffs.

Calculate the feature importance scores

For a fully connected NN, DeepPINK[14] used the multiplication of weight parameters from all layers as FIs. In DNNs with more complicated architecture, the multiplication of tensorial weights is not well-defined. To compute FIs in the proposed hierarchical DNN, we define FIs using the gradients of output $\mathbf{y} \in \mathbb{R}^n$ with respect to inputs $\mathbf{X} \in \mathbb{R}^{n \times p \times (M+1)}$,[32]. That is, the importance of feature $\mathbf{x}_j$ on the response $\mathbf{y} \in \mathbb{R}^n$ is measured by the local sensitivity of the predictive function to that feature. This is represented by a vector $\mathcal{T} = (\mathcal{T}_1, \mathcal{T}_2, \dots, \mathcal{T}_p)$, of length $p$, in which $\mathcal{T}_j = \mathbb{E}[\partial f(X)/\partial X_j]$ where $\mathbb{E}$ represents the expectation with respect to the joint distribution $(y, X_1, \dots, X_p)$ and $f$ represents the DNN. To compute this, we take the gradients for the input data $\mathbf{X} \in \mathbb{R}^{n \times p \times (M+1)}$, giving a gradient tensor $\mathbb{T} \in \mathbb{R}^{n \times p \times (M+1)}$, where $M$ is the number of knockoffs. Then, we take the average over samples which leads to the final feature importance matrix $\mathbf{T} \in \mathbb{R}^{p \times (M+1)}$. The $j$-th row of $\mathbf{T} \in \mathbb{R}^{p \times (M+1)}$ contains the feature importance scores of original and knockoffs for the $j$-th feature. Obtaining FIs with gradient information is architecture-independent; regardless of the neural network's architecture, the gradients of output with respect to inputs can be easily monitored and calculated.

Calculate the knockoff feature statistic

Assume $\mathbf{T} = [\mathbf{T}^0, \mathbf{T}^1, \dots, \mathbf{T}^M]$ is the matrix of FIs where $\mathbf{T}^0 \in \mathbb{R}^p$ is FIs for the original variants and the rest are for $M$ set of knockoffs. For the selection of important variants, the absolute values of feature importance scores (or absolute values of gradients) are passed to the knockoff selection procedure. For a single knockoff-based model, $W_j = |T_j^0| - |T_j^1|$. Intuitively, the original variants with higher FIs than its knockoffs are more likely to be causal. For multiple knockoffs, we used a multiple-knockoff feature statistics proposed by He et al.[15]. Two metrics $\kappa_j$ and $\tau_j$ are calculated for each feature $1 \leq j \leq p$ as follows:

$$\kappa_j = \arg\max_{0 \leq m \leq M} T_j^m, \quad 1 \leq j \leq p, \ m \in \{0, 1, \dots, M\}$$

$$\tau_j = T_j^{(0)} - \underset{1\leq m\leq M}{\text{median}}\, T_j^{(m)}$$

Where $\kappa_j$ denote the index of the original (denoted as 0) or knockoff feature that has the largest importance score; $\tau_j$ denote the difference between the largest importance score $T_j^{(0)}$ and the median of the remaining importance scores. The indexing in parenthesis refers to the ordered sequence of FIs in the descending order. Therefore, $T_j^{(0)}$ is the largest feature importance score for the *j*-th feature which can be either for the original feature or one of the knockoffs. The feature statistic is defined as

$$W_j = \left(T_j^0 - \underset{1\leq m\leq M}{\text{median}}\, T_j^m\right) I_{T_j^0 \geq \underset{1\leq m\leq M}{\max}\, T_j^m}$$

It has been empirically shown that the knockoff statistics with median substantially improves stability and reproducibility of knockoffs[15].

### Knockoff filter and Q-statistics for feature selection

The last step of knockoff framework is feature selection with FDR≤ $\alpha$, where $\alpha$ is a pre-defined bound for FDR, known as target FDR level. For a single knockoff, the feature statistic is defined as, $W_j = |T_j^0| - |T_j^1|$ and the knockoff threshold $\hat{t}$ is chosen as follows[11]:

$$\hat{t} = \min\left\{t > 0: \frac{1 + \#\{j: W_j \leq -t\}}{\#\{j: W_j \geq t\}} \leq \alpha\right\},$$

For multiple knockoffs,

$$\hat{t} = \min\left\{t > 0: \frac{\frac{1}{M} + \frac{1}{M}\#\{\kappa_j \geq 1, \tau_j \geq t\}}{\#\{\kappa_j = 0, \tau_j \geq t\}} \leq \alpha\right\}.$$

Variants with $W_j \geq \hat{t}$ are selected. Equivalently, a knockoff Q-value can be computed as

$$q_j = \min_{t \leq \tau_j} \frac{\frac{1}{M} + \frac{1}{M}\#\{j: \kappa_j \neq 0, \tau_j \geq t\}}{\#\{j: \kappa_j = 0, \tau_j \geq t\}}$$

for variants with statistics $\kappa = 0$, and $q_j = 1$ for variants with $\kappa \neq 0$. Selecting variants/windows with $W_j > \hat{t}$ is equivalent to selecting variants/windows with $q_j \leq \alpha$.

The advantage of the multiple-knockoff selection procedure is the new offset term $\frac{1}{M}$ (averaging over M knockoffs) that enables us to decrease the threshold of minimum number of rejections from $\left\lceil\frac{1}{\alpha}\right\rceil$ to $\left\lceil\frac{1}{M\alpha}\right\rceil$, leading to an improvement in the power. The use of median in the calculation of $W$ improves the stability.

**The proposed hierarchical deep learning structure**

To learn the complex nonlinear relationship among genetic variants, a fully connected neural network (FCNN) is utilized. Massive number of genetic variants $p$ can be deemed as a big computational burden. Additionally, to control FDR, the inclusion of the knockoff data which has the same size as the original cohort data adds more to both computational time and resources. Knowing the fact that the first layers of a deep learner include several weight parameters and it is essential to control the size of the neural network in its first layers, hierarchical deep neural networks are used to exponentially reduce the size of a DNN[33–36]. Assume that the number of neurons corresponds to each SNP is $(M + 1) \times p$ where $M$ is the number of knockoffs (in our experiments we set it to 5). In our proposed hierarchical deep learner, we group every original feature and its knockoffs in the first layer to a single neuron in the next layer through a nonlinear activation function. Hence, the size of neurons in the next layer reduces to $p$ neurons. We call this combination between two feature types as *feature-wise* hierarchy. Next, adjacent SNPs inherit similar traits and therefore one can take SNPs of the adjacent regions into the same groups. Assume that the number of SNPs in each group is set to $\sigma > 1$. Then every $\sigma$ neurons out of $p$ neurons in the second layer are grouped to a small set of $\theta$ neurons (channels) in the next layer. Hence, the number of neurons is further reduced to $\theta[p/\sigma]$. We call this combination between features in the second layer as *region-wise* hierarchy These are analogous to filters in a convolutional neural network. The architecture of the proposed HiDe-MK is illustrated in **Figure 1** and **Figure S1**. Also, see **Figure S2** for the impact of different number of kernel size on the observed FDR, power, and the total number of weight parameters of HiDe-MK. In our experiments, the number of channels in the second hierarchical layer is set to 8.

**Activation functions of HiDe-MK**

Deep learning consists of several layers in its structure which learn the underlying structure of data through nonlinear activation functions. Although sufficiently deep structure of neural network can learn complex features of real-world applications, having several layers in the DNN structure bring some challenges to the training such as vanishing gradient problem and the saturation problem of activation functions [37]. **Table 2** tabulates a list of important activation functions. The rectified linear unit (ReLU), which is shown by max{x,0}, is one of the most popular activation functions in deep learning[18] because of its outstanding performance and low computational cost compared to other activation functions such as the logistic sigmoid and the hyperbolic tangent[38]. However, if the data falls into the hard zero negative part of ReLU, many neurons will not be re-activated during the training process and its corresponding gradients are set to zero which avoid the weight update. This issue is known as the "dying ReLU" problem. In our experiments, ELU activation function showed best performance among other activation functions listed above.

**Table 2**. **The commonly used activation functions in the layers of DNN**. the underlying structure of data can be learned through linear or nonlinear activation functions.

| Activation function | Mathematical equation |
| --- | --- |
| ReLU | Max[z,0] |
| ELU | $f(z) = \begin{cases} z & if\ z > 0 \\ a(e^z - 1) & otherwise \end{cases}$ |
| Sigmoid | $(1 + e^{-z})^{-1}$ |
| Tanh | $(e^z - e^{-z})(e^z + e^{-z})^{-1}$ |
| Linear | Z |

**Training loss: MSE and BCE**

The mean squared error (MSE) is usually used as the loss function for training a *regression problem* as follows:

$$\mathcal{M}_{MSE}(\boldsymbol{y}, \hat{\boldsymbol{y}}) = \frac{1}{n} \sum_{i=1}^{n} (y_i - \hat{y}_i)^2,$$

where $\boldsymbol{y} \in \mathbb{R}^n$ represents the target value, $\hat{\boldsymbol{y}} \in \mathbb{R}^n$ represents the predicted value by the learner, and $n$ represents the total number of samples for every batch of data. We used cross-entropy loss, or log loss as loss function to measure the performance of a classification problem. Cross-entropy loss increases as the predicted probability of $\hat{\boldsymbol{y}} \in \mathbb{R}^n$ diverges from the actual label $\boldsymbol{y} \in \mathbb{R}^n$. For a binary classification problem, binary cross-entropy (BCE) is calculated as:

$$Loss_{BCE} = -(y \log(\hat{y}) + (1-y) \log(1-\hat{y}))$$

The output $\hat{\boldsymbol{y}} \in \mathbb{R}^n$ is treated as a probability value between 0 and 1. For a batch of data with $n$ data samples, the average cross-entropy is calculated as:

$$Loss_{Avg\_BCE} = -\frac{1}{n} \left[ \sum_{i=1}^{n} [y_i \log(\hat{y}_i) + (1-y_i) \log(1-\hat{y}_i)] \right]$$

In our training, we use the Adam optimizer to minimize the loss functions.

**Evaluation metrics: MSE and AUC**

The mean square error (MSE) is defined as mean of the square of the differences between actual values $\boldsymbol{y} \in \mathbb{R}^n$ and estimated values $\tilde{\boldsymbol{y}} \in \mathbb{R}^n$ which is same as we discussed in the previous section. The MSE is used as a metric for the regression task. One of the most popular evaluation metrics that is used to measure the ranking quality of a classifier is the area under an ROC curve (AUC). ROC is a probability curve of true positive rate and true negative rate and AUC represents the degree of separability between classes. For a binary classification task and a given classifier, let $\hat{x}_1, \ldots, \hat{x}_m$ be the output of the classifier on the positive examples and $\hat{y}_1, \ldots, \hat{y}_n$ be the output on the negative examples. Then, the AUC can be calculated as[39]:

$$AUC = \frac{\sum_{i=1}^{m} \sum_{j=1}^{n} 1_{\hat{x}_i > \hat{y}_j}}{mn}$$

where $1_X$ represents the indicator function of set $\mathbf{X} \in \mathbb{R}^{n \times p}$. This is a ranking method and a perfect ranking (i.e., $AUC = 1$) occurs if all positive examples are ranked greater than the negative ones.

**De-randomized feature importance values**

Interpretations of DNN methods are known to be fragile[9]. Due to the stochasticity of parameter initialization and optimization, different runs of DNNs including our proposed HiDe-MK may result in different sets of selected variants and hence different interpretations. Small changes in the weight or gradient values may change the set of selected variants chosen by a knockoff statistic in addition to slight changes in terms of accuracy between different runs. We empirically observed different runs give different set of feature importance values and different set of selected variants. Intuitively, most strong signals can be detected at low target FDR threshold value, and they almost appear in every run, but there are weak signals that make a difference for two different runs. A more consistent FI score is desirable to minimize the false positive rate. Therefore, we devise a de-randomization mechanism to stabilize our HiDe-MK learner. We first conducted 5-fold cross validation to select the learning parameters that exhibit highest validation AUC at an optimal epoch number. We chose the optimal epoch number such that the validation AUC is maximized and stable through $|Loss_k - Loss_{min}|/|Loss_{min}| < 0.01$ for a range from $k$ - 5 to $k$ + 5 where $k$ is the index for the epoch number. **Figure S3** show the optimal epoch number through the validation loss for real data analysis. Then, we repeatedly fit our final model using the whole dataset with optimal hyper-parameters and the optimal epoch number for ten times and take the median of feature importance scores (FIs). The knockoff statistic is then applied to the median of FIs. With this approach, we observed improved power and controlled FDR. Results of de-randomized HiDe-MK is illustrated in **Figure. 2**.

**DNN configurations**

We search for optimal hyper-parameters of HiDe-MK through a randomized search with a 5-fold cross validation. In our search space for simulation data, we have 25 sets of hyper-parameters that adjust L1-norm coefficient through {1e-5, 5e-5, …,1e-3} and epoch number in the range 25 to 80 with step size 5. We do this for every replicate of 200 replicates and for both dichotomous and quantitative traits. We used batch size 1024, a base learning rate 0.001, the number of filters 8, and the kernel size is set to 5. The optimal number of epochs is obtained through the maximum validation loss and then is applied to the whole data and FIs are only stored for the optimal epoch number. Then, we apply knockoff statistics and measure FDR and Power. For the real data analysis, we trained HiDe-MK up to 250 epochs which lasts about 12h applied to the imputed data. The optimal number of epochs for the real data analysis is 92. **Figure S3** illustrates this through the validation loss for real data analysis. Based on a randomized search and 5-fold cross validation, we obtained the optimal hyper-parameters with learning rate 0.00807, L1-norm coefficient was 0.00026, kernel size for the second layer of hierarchy 1800. The number of filters was 8. The number of neurons in the dense layer was set to 50. With the optimal set of hyper-parameters and the network architecture, the final size of HiDe-MK or the total number of weight parameters was 204,925. We only used a L-1 norm regularizer for the first locally connected layer. The activation function was 'ELU' for every layer except the last layer. For the decision layer, we used Sigmoid activation function for dichotomous trait and linear activation function for the continuous trait in simulation studies and real data analysis. To do experiments, we used Sherlock high-performance computing (HPC) device at Stanford University. The configurations for cluster of remote machines were GPU: Nvidia GeForce RTX 2080 Ti 11GB; and CPU: Intel Xeon Silver 4116 2.10GHz; OS: Ubuntu16.04.3 LTS.

**Data availability**

Alzheimer's disease genetic cohort data can be obtained for approved research, see description in Le Guen et al. (2021)[31].

**Code Availability**

The code for the generation and reproduction of the simulation studies of SKAT haplotype data is written in R programming language. The code for HiDe-MK training, prediction and evaluation were written in Python with Keras and Tensorflow. The codes are feely available at: https://github.com/Peyman-HK/De-randomized-HiDe-MK MK.

**Acknowledgments**

This research was supported by NIH/NIA award AG066206 (ZH).

**Author Contributions**

P.H.K, and Z.H. developed the concepts for the manuscript and proposed the method. P.H.K, F.L., Y.L.G., and Z.H. designed the analyses and applications and discussed results. P.H.K, Z.H., and F.L. conducted the analyses. Z.H. and Y.L.G., helped interpret the results of the real data analyses. P.H.K., Z.H., F.L. and Y.L.G. prepared the manuscript and contributed to editing the paper.

**Competing Interests**

The authors declare no competing interests.

## Supplementary Materials

**Figure S1. Architecture of HiDe-MK for real data analysis.** The framework is applied to a data with 11,662 genetic variants with five sets of knockoffs; every knockoff cohort has the same size as the original cohort. The covariates with 11 features are fed to the neural network right before the last layer. The genotype data is presented to the network as the first input data with size $11662 \times 6$ where 6 is number of original and knockoff features. The term "?" is referred to the batch size, and in our experiments, we used 1024 as the batch size. In the next layer, each 6 neurons, i.e., one original feature and its 5 knockoffs are grouped together. The shape of matrix changes from $11662 \times 6$ to $11662 \times 1$. Next, every 1800 neurons are grouped and are further mapped to 8 channels. Doing so, the shape of data changes from $11662 \times 1$ to $6 \times 8$. Then two-dimensional data is flattened from the size $6 \times 8$ to $48 \times 1$. These 48 neurons are passed to a dense layer with 50 neurons. The 11 covariates are then concatenated to these 50 neurons. Note that the activation function for covariate layer is linear. The last layer with one neuron is the decision layer. This whole process is repeated for the whole samples of data till the total number of epochs are completed.

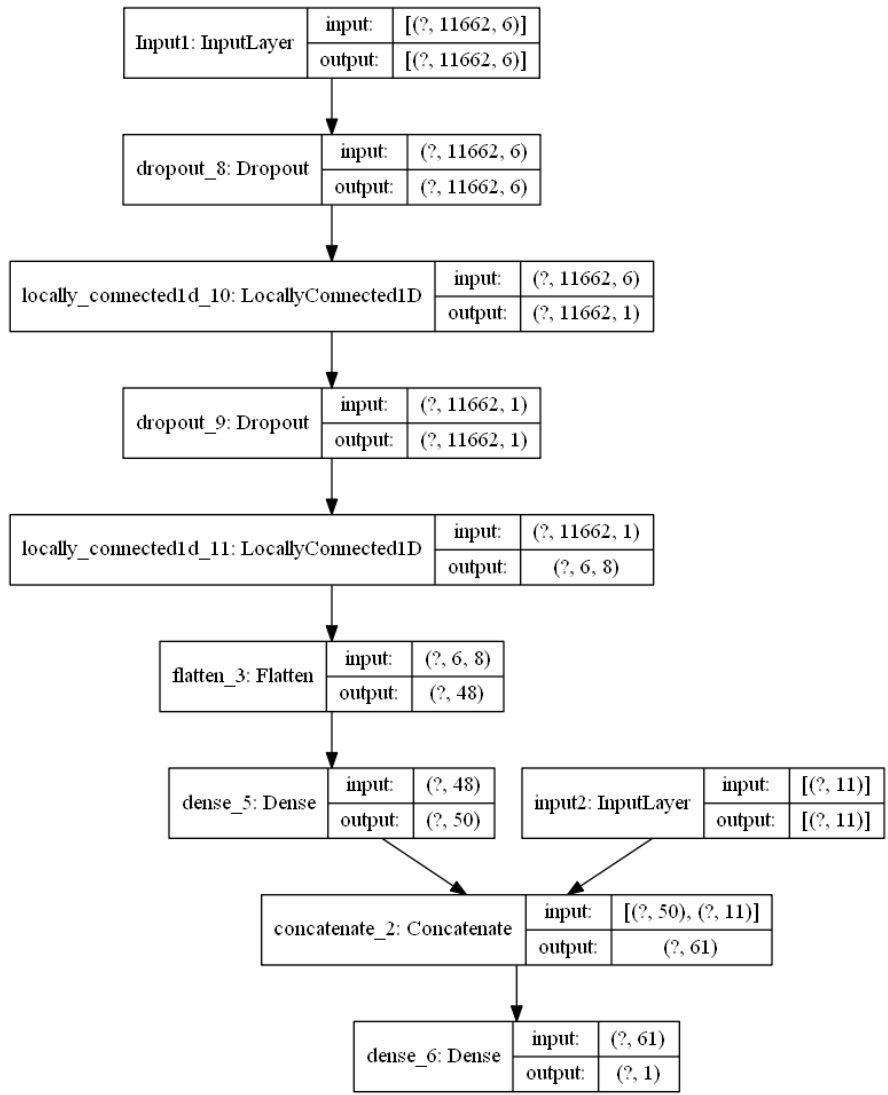

**Figure S2. Impact of different kernel size values on measured FDR, power, and the total number of weight parameters**. The effectiveness of the locally connected layers in the architecture of HiDe-MK depends on the proper choice of the kernel size, i.e., the number of neurons that should be grouped. To empirically validate its importance on HiDe-MK performance, we ran a HiDe-MK on dichotomous trait with different kernel size values and measured the average FDR, power, and the size of network over 50 runs. The search space for other hyperparameters of HiDe-MK is same as we discussed in Results' section. Kernel size 25 has superior power while FDR is under control and the number of weight parameters are much less than when it is compared to the small kernel size 5.

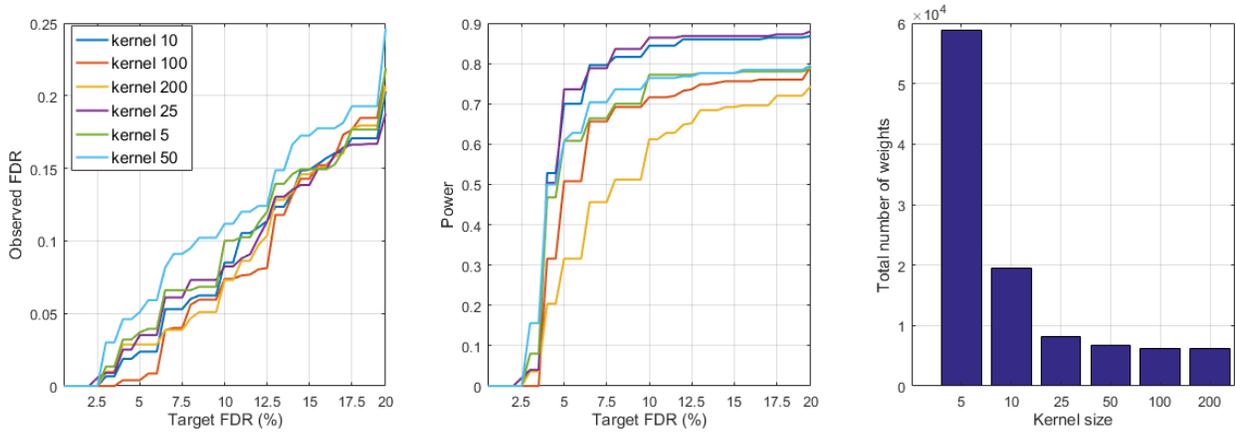

**Figure S3. The choice of optimal epoch number through monitoring validation AUC and validation loss on real data analysis**. The convergence of HiDe-MK can be monitored via the validation loss and validation AUC. This also helps to choose the optimal epoch number. The block hollow circle displays where HiDe-MK reaches its optimal validation loss. The y-axis is shown in log scale which reveals HiDe-MK is slightly overfitted as the number of epochs keeps increasing.

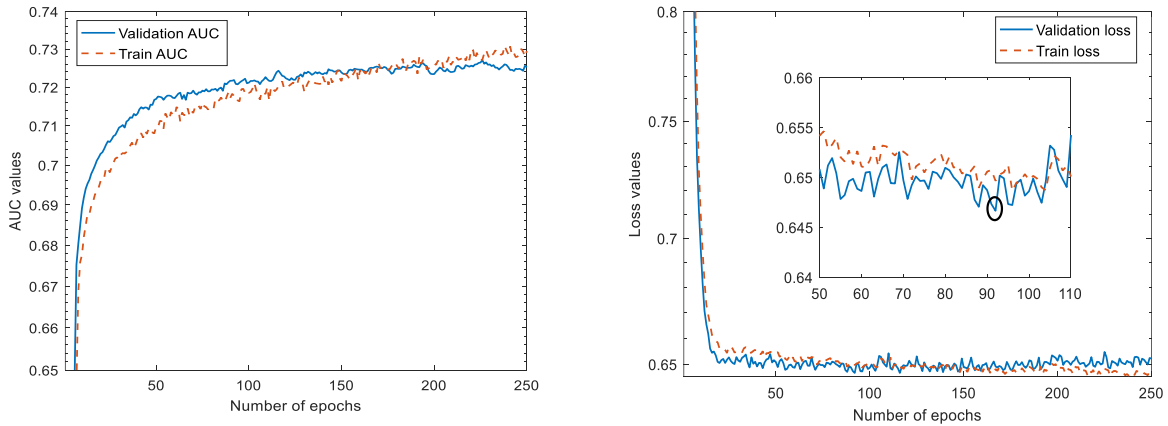

**Figure S4. De-randomized HiDe-MK demonstrates low randomness compared to a single run of HiDe-MK in terms of averaging.** The correlation matrix of ten different runs of HiDe-MK with identical learning hyper-parameters reveals a drastic randomness in its every run (Left). Taking average of feature importance values (FIs) for every ten runs substantially de-randomized FIs (Right). In the Results section we displayed similar plots in terms of median.

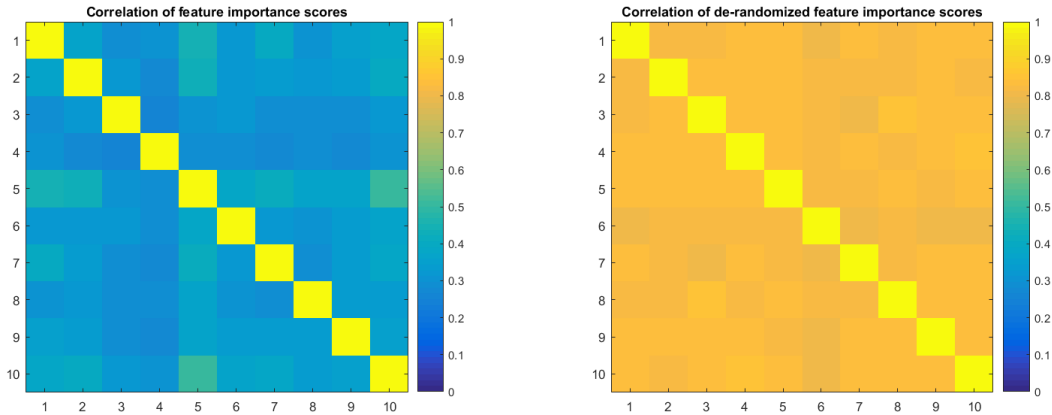

**Table S1. Genetic markers identified by a de-randomized HiDe-MK (hg38).**

| # | Position(hg38) | Variant (frequency) | rs-IDs | Gene |
|---|---|---|---|---|
| 1 | chr2:127072367 | C (19.5807%) | rs3754609 | *BIN1* |
| 2 | chr2:127086585 | A (35.8064%) | rs10929006 | *BIN1* |
| 3 | chr2:127135234 | T (39.0325%) | rs6733839 | (*BIN1, CYP27C*) |
| 4 | chr3:57873891 | T (18.3600%) | rs141780045 | *SLMAP* |
| 5 | chr4:11021175 | G (27.1969%) | rs4373155 | (*MIR572, HS3TST1*) |
| 6 | chr6:41187262 | rs2093396 | A (30.6354%) | (*TREM2, TREML2*) |
| 7 | chr6:41187288 | C (30.6847%) | rs2093395 | (*TREM2, TREML2*) |
| 8 | chr6:47459545 | A (23.6902%) | rs1931833 | (*CD2AP-DT, CD2AP*) |
| 9 | chr6:47515917 | G (23.3414%) | rs4715025 | *CD2AP* |
| 10 | chr6:47517266 | C (24.2702%) | rs7749271 | *CD2AP* |
| 11 | chr6:47520026 | G (23.3376%) | rs10948363 | *CD2AP* |
| 12 | chr6:47521368 | G (22.1927%) | rs6913202 | *CD2AP* |
| 13 | chr7:54852014 | G (9.6292%) | rs117319336 | (*SEC61G, SEC61G-DT*) |
| 14 | chr7:100171334 | T (10.0948%) | rs75779608 | *GPC2* |
| 15 | chr7:100172833 | T (21.9084%) | rs57966707 | *GPC2* |
| 16 | chr7:143402014 | G (18.7998%) | rs7791765 | *EPHA1* |
| 17 | chr7:143412115 | G (15.3423%) | rs75045569 | *EPHA1-AS1* |

| 18 | chr8:144103704 | A (1.6911%) | rs34173062 | *SHARPIN* |
| --- | --- | --- | --- | --- |
| 19 | chr11:47363046 | G (32.5082%) | rs11039203 | *SPI1* |
| 20 | chr11:86043772 | C (11.7067%) | rs17817931 | *PICALM* |
| 21 | chr11:86067709 | T (11.2518%) | rs17817992 | *PICALM* |
| 22 | chr14:92460608 | T (18.5875%) | rs10498633 | *SLC24A4* |
| 23 | chr19:44881148 | C (4.7236%) | rs73052307 | *NECTIN2* |
| 24 | chr19:44885243 | G (20.3541%) | rs283811 | *NECTIN2* |
| 25 | chr19:44887076 | G (22.6363%) | rs283815 | *NECTIN2* |
| 26 | chr19:44888997 | T (12.9502%) | rs6857 | *NECTIN2* |
| 27 | chr19:44891079 | C (12.3285%) | rs71352238 | *TOMM40* |
| 28 | chr19:44891712 | G (8.9215%) | rs184017 | *TOMM40* |
| 29 | chr19:44892362 | G (12.5052%) | rs2075650 | *TOMM40* |
| 30 | chr19:44892457 | C (23.3946%) | rs157581 | *TOMM40* |
| 31 | chr19:44892652 | G (12.8744%) | rs34404554 | *TOMM40* |
| 32 | chr19:44892962 | T (22.1318%) | rs157582 | *TOMM40* |
| 33 | chr19:44892962-44892964 | DELETION (0.0026%) | rs778934950 | *TOMM40* |
| 34 | chr19:44893408 | T (21.9918%) | rs59007384 | *TOMM40* |
| 35 | chr19:44894695 | C (2.3694%) | rs116881820 | *TOMM40* |
| 36 | chr19:44895376 | C (12.2640%) | rs11668327 | *TOMM40* |
| 37 | chr19:44897468 | T (2.3542%) | rs114536010 | *TOMM40* |
| 38 | chr19:44899959 | T (2.4376%) | rs115881343 | *TOMM40* |
| 39 | chr19:44901600 | C (2.2708%) | rs112019714 | *TOMM40* |
| 40 | chr19:44901805 | G (45.0679%) | rs1038026 | *TOMM40* |
| 41 | chr19:44903281 | CG (2.0737%) | - | *TOMM40* |
| 42 | chr19:44904531 | A (43.2103%) | rs7259620 | *(TOMM40, APOE)* |
| 43 | chr19:44905307 | T (16.4683%) | rs449647 | *(TOMM40, APOE)* |
| 44 | chr19:44906745 | A (11.5665%) | rs769449 | *APOE* |
| 45 | chr19:44908684 | C (9.3349%) | rs429358 | *APOE* |
| 46 | chr19:44908822 | T (3.3493%) | rs7412 | *APOE* |
| 47 | chr19:44909698 | C (2.4755%) | rs1081105 | *AC011481.3* |
| 48 | chr19:44909976 | T (3.4461%) | rs1065853 | *AC011481.3* |
| 49 | chr19:44911142 | A (9.9970%) | rs72654473 | *AC011481.3* |
| 50 | chr19:44912383 | A (11.6347%) | rs445925 | *AC011481.3* |
| 51 | chr19:44912456 | A (10.9144%) | rs10414043 | *AC011481.3* |
| 52 | chr19:44912678 | T (10.8082%) | rs7256200 | *AC011481.3* |
| 53 | chr19:44912921 | T (23.0002%) | rs483082 | *AC011481.3* |
| 54 | chr19:44913484 | T (22.0411%) | rs438811 | *AC011481.3* |
| 55 | chr19:44913574 | G (11.2670%) | rs390082 | *AC011481.3* |
| 56 | chr19:44915533 | C (18.2993%) | rs5117 | *APOC1* |
| 57 | chr19:44921921 | A (2.6158%) | rs190712692 | *(AC011481.3, APOC1)* |
| 58 | chr19:44923535 | A (5.7320%) | rs141622900 | *(APOC1, APOC1P1)* |
| 59 | chr19:44923868 | A (11.2063%) | rs111789331 | *(APOC1, APOC1P1)* |
| 60 | chr19:44924977 | A (12.1313%) | rs66626994 | *(APOC1, APOC1P1)* |
| 61 | chr19:44973974 | G (16.7033%) | rs57465754 | *CLPTM1* |
| 62 | chr19:44916825 | C (2.1874%) | rs73052335 | *APOC1* |
| 63 | chr19:44917947 | T (2.5590%) | rs150966173 | *APOC1* |
| 64 | chr19:44917997 | A (11.1229%) | rs12721046 | *APOC1* |
| 65 | chr19:44918393 | A (1.6074%) | rs140480140 | *APOC1* |
| 66 | chr19:44918903 | G (6.1776%) | rs12721051 | *APOC1* |
| 67 | chr19:44919304 | G (4.0147%) | rs1064725 | *APOC1* |
| 68 | chr19:44919589 | A (16.0854%) | rs56131196 | *(AC011481.3, APOC1)* |
| 69 | chr19:44919689 | G (16.1119%) | rs4420638 | *(APOE, APOC1)* |
| 70 | chr19:44921809 | A (0.7696%) | rs188535946 | *(AC011481.3, APOC1)* |
| 71 | chr19:44921915-44921921 | ATGGGCG (0.0038%) | - | - |
| 72 | chr19:44921918-44921921 | GGCG (1.3382%) | - | - |